\documentclass[journal]{IEEEtran}
\ifCLASSINFOpdf
\else
\fi

\usepackage{times}
\usepackage{helvet}
\usepackage{courier}
\usepackage{algorithm}
\usepackage{algorithmic}
\usepackage{amsthm}
\usepackage{amsmath}
\usepackage{amssymb}
\usepackage{bm}
\usepackage[mathscr]{euscript}
\usepackage{multirow}
\usepackage{graphics}
\usepackage{graphicx}
\usepackage{caption}
\usepackage{subcaption}
\newtheorem{theorem}{Theorem}

\newtheorem{definition}[theorem]{Definition}

\newcommand{\matr}[1]{\bm{#1}}

\usepackage{array}
\newcolumntype{C}{>{\centering\arraybackslash}m{1.42cm}}
\newcolumntype{D}{>{\centering\arraybackslash}m{0.77cm}}

\begin{document}
%
\title{Magnitude Bounded Matrix Factorisation \\ for Recommender Systems}
%
%
%

\author{Shuai Jiang, Kan Li, and Richard Yi Da Xu
	\thanks{S. Jiang is with the School of Computer Science, Beijing Institute of Technology (BIT), Beijing, China, and the Faculty of Engineering and Information Technology, University of Technology Sydney (UTS), Australia (email: jiangshuai@bit.edu.cn; shuai.jiang-1@student.uts.edu.au).}
    \thanks{K. Li is with the School of Computer Science, Beijing Institute of Technology (BIT), Beijing, China (email: likan@bit.edu.cn).}
    \thanks{R. Y. D. Xu is with the Faculty of Engineering and Information Technology, University of Technology Sydney (UTS), Australia (email: yida.xu@uts.edu.au).}
}

%
%

\markboth{IEEE TRANSACTIONS ON NEURAL NETWORKS AND LEARNING SYSTEMS}%
{Shell \MakeLowercase{\textit{et al.}}: Bare Demo of IEEEtran.cls for IEEE Journals}
%



\maketitle

\begin{abstract}
	Low rank matrix factorisation is often used in recommender systems as a way of extracting latent features. When dealing with large and sparse datasets, traditional recommendation algorithms face the problem of acquiring large, unrestrained, fluctuating values over predictions especially for users/items with very few corresponding observations. Although the problem has been somewhat solved by imposing bounding constraints over its objectives, and/or over all entries to be within a fixed range, in terms of gaining better recommendations, these approaches have two major shortcomings that we aim to mitigate in this work: one is they can only deal with one pair of fixed bounds for all entries, and the other one is they are very time-consuming when applied on large scale recommender systems. In this paper, we propose a novel algorithm named Magnitude Bounded Matrix Factorisation (MBMF), which allows different bounds for individual users/items and performs very fast on large scale datasets. The key idea of our algorithm is to construct a model by constraining the magnitudes of each individual user/item feature vector. We achieve this by converting from the Cartesian to Spherical coordinate system with radii set as the corresponding magnitudes, which allows the above constrained optimisation problem to become an unconstrained one. The Stochastic Gradient Descent (SGD) method is then applied to solve the unconstrained task efficiently. Experiments on synthetic and real datasets demonstrate that in most cases the proposed MBMF is superior over all existing algorithms in terms of accuracy and time complexity.
\end{abstract}


%
\IEEEpeerreviewmaketitle

\section{Introduction}
%
%
%
%
\IEEEPARstart{T}{he} prevalence of recommender systems has become evident over recent years. Amongst many of its algorithms, Collaborative Filtering (CF) has been one of the most widely used \cite{ZhengAAAI2010,WangKDD2011,BobadillaKBS2013}. The two primary categories of CF are the neighbourhood methods and latent factor models. The former computes the relationships between users or items \cite{BellICDM2007,KorenKDD2008}, while the latter tries to explain the ratings by characterising both users and items on factors inferred from the ratings patterns \cite{GuICDM2010,BaltrunasRS2011}, which has become more popular in recent years. The most successful latent factor models used low rank Matrix factorisation (MF) to obtain the feature vectors, in which the goal is to fill the unknown entries of the data matrix by calculating the inner product of factorised feature matrices \cite{RennieICML2005,LuoKBS2013}. After obtaining the prediction values, recommendations on items can be made by ranking the values associated with the user in question. 

Although existing MF-based CF algorithms have been proved effective for general recommender systems, such as Movielens \cite{HarperTIIS2015} and Jester \cite{GoldbergIR2001} datasets, when it comes down to datasets with high sparsity where many users only provide very few ratings, such as BookCrossing \cite{ZieglerWWW2005}, Dating \cite{Brozovsky2007arXiv} and Netflix Prize \cite{Netflix} datasets, the predictions of missing entries are often unstable and out of range \cite{KorenC2009}. 

\begin{figure}[!t]
	\begin{subfigure}[t]{0.47\linewidth}
		\begin{center}
			\includegraphics[width=\textwidth]{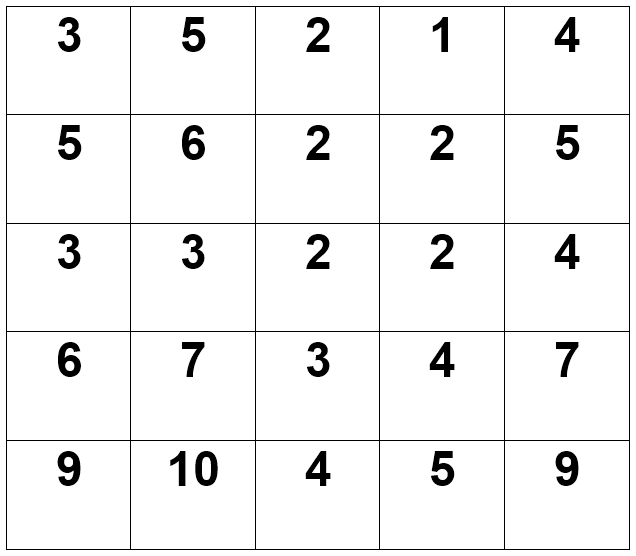}
			\caption{complete observation matrix}
		\end{center}
	\end{subfigure}
	\hspace{2ex}
	\begin{subfigure}[t]{0.47\linewidth}
		\begin{center}
			\includegraphics[width=\textwidth]{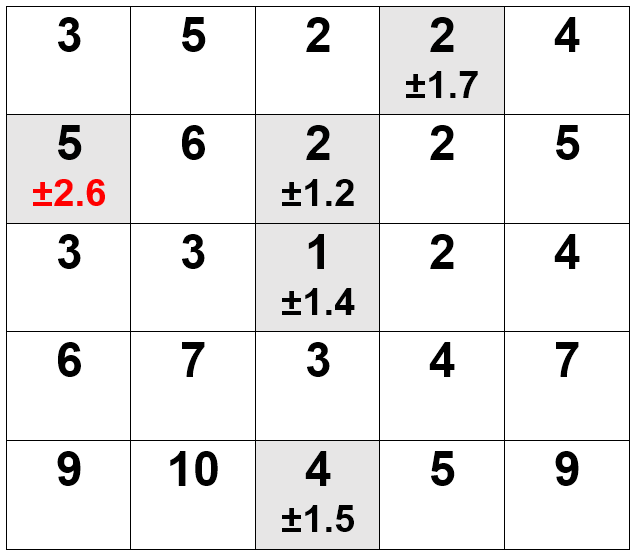}
			\caption{$d = 0.8$, $\text{ave}(\sigma) = 1.68$, $\max(\sigma) = 2.6$}
		\end{center}
	\end{subfigure}	
	\par\medskip
	\begin{subfigure}[t]{0.47\linewidth}
		\begin{center}
			\includegraphics[width=\textwidth]{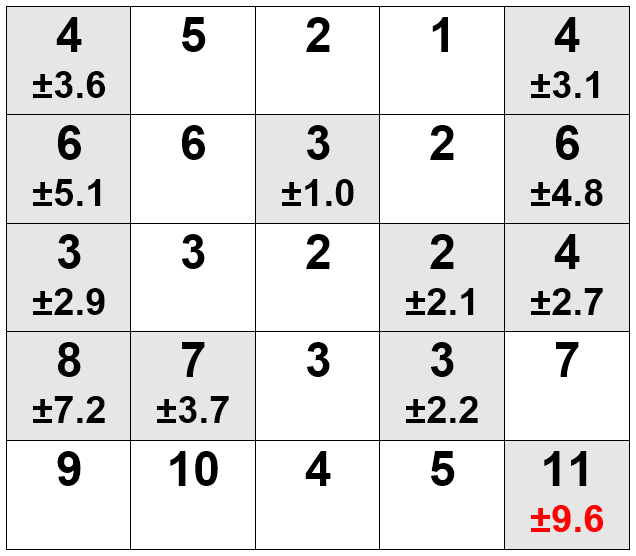}
			\caption{$d = 0.5$, $\text{ave}(\sigma) = 4.00$, $\max(\sigma) = 9.6$}
		\end{center}
	\end{subfigure}
	\hspace{2ex}
	\begin{subfigure}[t]{0.47\linewidth}
		\begin{center}
			\includegraphics[width=\textwidth]{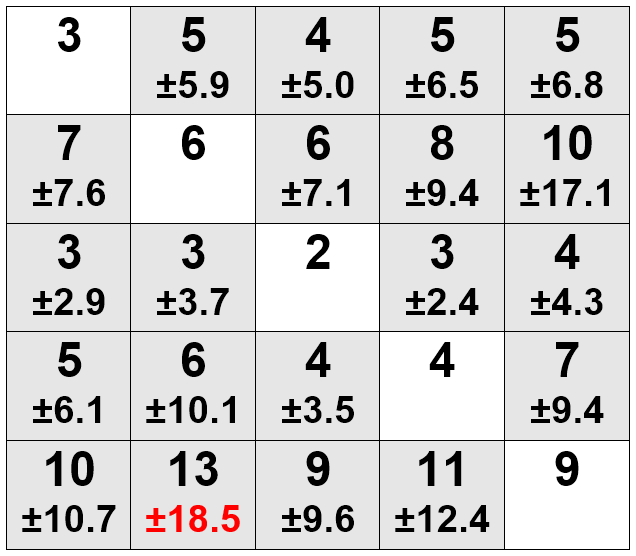}
			\caption{$d = 0.2$, $\text{ave}(\sigma) = 7.95$, $\max(\sigma) = 18.5$}
		\end{center}
	\end{subfigure}
	\caption{Predictions of Non-negative Matrix Factorisation (NMF) pertaining to different matrix densities. Numbers in grey shadows are the average recovered entries after $1,000$ repetitions, along with the standard deviation presented next to $\pm$. The density is denoted by $d$ while $\text{ave}(\sigma)$ and $\max(\sigma)$ respectively represent the average and maximal standard deviation of the recovered entries. \label{eg1}}
\end{figure}

We demonstrate this effect through an example of a synthetic recommender system in Figure \ref{eg1}. In this example, we randomly removed parts of entries from the original complete observation matrix (Figure \ref{eg1} (a)) with different remaining proportions (Figure \ref{eg1} (b), (c) and (d)). Then we ran the NMF method \cite{LeeNIPS2001} on each matrix $1,000$ times to calculate the average predictions (in grey cells) as well as its average and maximal standard deviations ($\sigma$). As seen from the results, the increase in missing entries led towards higher fluctuations as both the average and maximal $\sigma$ became greater. One solution would be to repeat the algorithm many times to reduce the predication variance. However, this will cause an increase in computational cost, especially when large data sets are involved.

Reducing fluctuation on predictions can help improve recommendation accuracy on large and sparse datasets. Since many recommendation systems have a single predefined range for all entries, one straightforward way to reduce fluctuation over predictions is to limit the product of the factorised matrices, to make them fall within a predefined range. There are two types of algorithms in literature seeking to address this bounding problem: one is to project the products that are out of range to the boundaries \cite{KannanKIS2014,FangIJCAI2017}, the other one is to impose strong penalties in the objective function if the products are out of range \cite{LeICEITES2015}. Projecting (or mapping) methods require in range or out of range checks on each of the product entries obtained by the multiplication of factorised matrices, which causes the issue of high computation time especially on large scale dataset (see the Section of Complexity Analysis for more details), while the performance of the penalty methods highly depends on the setting of their penalty coefficients.

Apart from the above mentioned recommendation systems with predefined single range, there are many other ones where data does not share a single range: for instance, in ``user-location check-in frequency'' dataset in \cite{ChengAAAI2012}, the matrix contains the frequency counts of users appearing in different locations, where the scales of frequencies vary from user to user and location to location. In these cases, the above two bounding methods may fail because they can only address datasets with single range: imposing a single pair of bounds, for example, using the maximum and minimum of all known values, will have very little effect on bounding most of the entries, as they are all far from the end points of this range. From another aspect, even when dealing with datasets which have a single predefined range, such as movie ratings in Movielens which are between $1$ and $5$, bounding the entries in the product matrix into different and more concise ranges (e.g. related to the corresponding user or item) is also likely to gain better recommendations. 

In this paper, we propose a novel model called, Magnitude Bounded Matrix Factorisation (MBMF), which is a bounding model designed for large and sparse recommender systems. As other bounding approaches, MBMF deals with the fluctuation of recommendations by bounding the entries in the product matrix into predefined ranges. The differences are, MBMF can be applied to datasets with or without predefined single bounds, and allows the product entries to have individually different bounds. In the proposed MBMF, we first devise an optimisation algorithm where individual range constraints are placed onto the magnitude of latent feature vectors. This allows MBMF to deal with any datasets for which the magnitudes can be set using context-specific information. Secondly, in order to solve this constrained optimisation problem, we convert the original constrained problem into an unconstrained one by using coordinate conversion such that the problem becomes that of the unconstrained task. Then, we apply a stochastic gradient descent approach to solve the unconstrained optimisation efficiently.

The main contributions of this paper are:

1. We propose a novel algorithm named MBMF, a bounding model which deals with the fluctuation of recommendations on large and sparse datasets with or without predefined single bounds. Different from existing studies, MBMF bounds all entries using individual ranges. To the best of our knowledge, MBMF is the first bounding model which allows different bounds for each entry based on MF framework for recommender systems.

2. We found MBMF is the fastest bounding algorithm against the state-of-art bounding algorithms. On very large synthetic datasets (e.g. $200,000$ by $500,000$), MBMF is the only bounding algorithm which has comparable efficiency against classic matrix factorisation algorithms (MF and NMF).

3. We introduce a feasible way to incorporate historical data into our proposed model, which considers both the global influence and the individual effect. Experiments on four real world recommendation systems show that MBMF is capable for different types of datasets and can achieve the best performance in most cases.

The rest of this paper is organised as follows: in the next section, we review related literature to our work. The Method section contains the details of our algorithm, followed by the Experiments section in which we evaluate our algorithm as well as the Conclusions.

\section{Related Work}

For recommender systems, MF algorithms are mostly used in the latent factor models, where the features of users/items are represented by the factorised matrices. Many variations have been proposed based on the MF framework, with different additional information, such as user/item bias \cite{KorenC2009,ZhengAAAI2010}, meta information collected from users and items \cite{GuICDM2010}, and temporal dynamics \cite{KorenCACM2010,ZhengWWW2010}. 

As for studies on the bounds of product matrix, Bounded Matrix Factorisation (BMF) proposed by \cite{KannanKIS2014} was the first one focusing on limitations of ranges over predictive values in recommender systems. The key thought of BMF is the following updating formula:
\begin{equation}
\matr{W}_{ab} = 
\begin{cases}
\max(\textbf{l}), \quad if \quad \matr{W}_{ab}^{*} < \max(\textbf{l}) \\
\min(\textbf{u}), \quad if \quad \matr{W}_{ab}^{*} > \min(\textbf{u}) \\
\matr{W}_{ab}^{*}, \quad otherwise,
\end{cases}
\end{equation}
where $\textbf{l}$ is the lower bound vector, $\textbf{u}$ is the upper bound vector, $\matr{W}_{ab}$ is the element in question and $\matr{W}_{ab}^{*}$ is the optimal solution derived from the objective function by letting the derivatives vanish. Under the above updating rules, BMF is proved to achieve better performance than unbounded methods. However, BMF has been proven not to converge to stationary points \cite{FangIJCAI2017}. Apart from the convergence, its core idea is similar to a truncated approach: instead of directly truncating entries to the predefined bounds, BMF moves the updating values to the inner bounds (the minimum of the upper bound vector and the maximum of the lower bound vector) if they are outside of the given range. As for the time complexity, BMF has to compute for each entry a complementary matrix whose size is close to the observation matrix, which can be time consuming when big datasets are involved.

Recently, an improved bounded matrix completion method was proposed in \cite{FangIJCAI2017} based on Alternating Direction of Multiplier Method (ADMM) framework (so is called BMC-ADMM). The optimisation problem in their work is defined as:
\begin{equation}
\begin{aligned}
\min_{\matr{X},\matr{Z},\matr{T}} & \frac{1}{2} \lVert \matr{P} \ast (\matr{V} - \matr{X}) \rVert_{F}^{2} + \lambda \lVert \matr{Z} \rVert_{\ast} \\
s.t. & \matr{X} + \matr{E} = Z, \neg \matr{P} \ast \matr{X} = \matr{0}, \matr{P} \ast \matr{E} = 0, \\
& \matr{Z} = \matr{T}, r_{\text{min}} \leqslant \matr{T} \leqslant r_{\text{max}},
\end{aligned}
\end{equation}
where $\lVert \cdot \rVert_{F}$ is the Frobenius norm, $\matr{P}$ is a mask matrix, $\ast$ denotes the element-wise multiplication, $\lVert \cdot \rVert_{\ast}$ is the nuclear norm commonly used to approximate the matrix rank, and $\neg$ denotes logical NOT. The authors aimed to improve the theoretical foundation in BMF, and to reformulate the model so that it can take advantage of the ADMM framework (split the loss, regularisation and constraints) which guarantees their algorithm to converge to the optimal solution. However, the paper does not differ fundamentally from BMF as far as bounding the predicted entries is concerned: during each iteration, BMF-ADMM checks the product of the two factorised matrices and projects any entries which are outside of the range to the boundaries. This checking mechanism is also the most time consuming step in the model which was reported in their work. Another potential obstacle for BMC-ADMM to be applied on very large datasets is the fact that, the updating procedure involves the calculation of partial singular value decomposition (SVD) for a non-sparse matrix having the same size of the observation matrix. Although the authors used several tricks to accelerate each of the updating steps, its efficiency still falls behind of the classic methods and our proposed MBMF (see Complexity Analysis for more details). 

An alternative way to bound the entries is the so-called, Bounded-SVD \cite{LeICEITES2015}. In their work, the authors imposed a penalty regularisation term to penalise the updating values which tend to fall outside of the given range:
\begin{equation}
\begin{aligned}
F & = \sum_{ij} (\matr{V}_{ij} - \matr{W}_{i:}\matr{H}_{:j}) + \\
&\quad \lambda(e^{\alpha(\matr{W}_{i:}\matr{H}_{:j} - r_{\text{max}})} + e^{\alpha(r_{\text{min}} - \matr{W}_{i:}\matr{H}_{:j})}),
\end{aligned}
\end{equation}
where $\lambda$ and $\alpha$ are parameters controlling the strength of penalties, $r_{\text{max}}$ and $r_{\text{min}}$ are the given upper and lower bounds. This regularisation method is complex, and cannot guarantee all product values are bounded within the given range simultaneously: therefore, the objective function only considers a small set of entries instead of bounding all of them. Besides, from our experiments, Bounded-SVD is very sensitive to its penalty coefficients: when they are set to be large (such as $10$), all product values become very small, while when they are set to have small values (such as $0.01$), most of the entries cannot be bounded within the predefined range. This algorithm also faces a time complexity issue when applied on large datasets.

MBMF is different from all the above existing bounding algorithms. It neither shifts the product values to fall between the boundaries, nor imposes any penalties in the objective function. MBMF introduces the magnitude constraints which guarantee the product of factorised matrices to be within ranges determined by these magnitude constraints. With the help of coordinates conversion, it manages to model the problem with an unconstrained optimisation task which can be solved efficiently by the well-known SGD method.

\section{Method}

In this section, we introduce a novel matrix factorisation algorithm called Magnitude Bounded Matrix Factorisation (MBMF) to deal with the prediction fluctuation for users/item in recommender systems. We first define the magnitude bounded matrix factorisation problem, followed by conversion from the constrained task to an unconstrained one. After that, we apply a stochastic gradient descent method to solve the unconstrained task.

\subsection{Magnitude Bounded Matrix Factorisation Problem}

Given an observation matrix $\matr{V} \in \mathbb{R}^{N \times M}$, the original low rank matrix factorisation task is to find two factor matrices $\matr{W} \in \mathbb{R}^{N \times K}$ and $\matr{H} \in \mathbb{R}^{K \times M}$ which satisfy the approximation $\matr{V} \approx \matr{W} \matr{H}$, where $K$ is commonly referred to as the latent dimension and usually set a smaller value than $N$ and $M$. According to the definition of matrix multiplication and the geometric interpretation of vector inner product in Euclidean space, each element of $\matr{V}$, say $\matr{V}_{ij}$, is approximated by the inner product of the $i^{\text{th}}$ row of $\matr{W}$ and the $j^{\text{th}}$ column of $\matr{H}$:
\begin{equation}
\matr{V}_{ij} \approx \matr{\hat{V}}_{ij} = \matr{W}_{i:} \cdot \matr{H}_{:j} = |\matr{W}_{i:}| |\matr{H}_{:j}| \cos\alpha_{ij}
\end{equation}
where $\matr{W}_{i:}$ denotes the $i^{\text{th}}$ row of $\matr{W}$ (so does $\matr{H}_{:j}$), $|\cdot|$ denotes the vector magnitude and $\alpha_{ij}$ is the angle between vectors $\matr{W}_{i:}$ and $\matr{H}_{:j}$. As $\cos\alpha_{ij} \in [-1,1]$, the prediction entry $\matr{\hat{V}}_{ij}$ is bounded by the product of magnitudes:
\begin{equation}
-|\matr{W}_{i:}| |\matr{H}_{:j}| \leqslant \matr{\hat{V}}_{ij} \leqslant |\matr{W}_{i:}| |\matr{H}_{:j}|.
\end{equation}

\begin{definition} [Magnitude Bounded Matrix Factorisation Problem]
	Given an observation matrix $\matr{V} \in \mathbb{R}^{N \times M}$ and two constant positive magnitude vectors $\bm{r}^{\matr{W}} \in \mathbb{R}^{N}_{+}$ and $\bm{r}^{\matr{H}} \in \mathbb{R}^{M}_{+}$, magnitude bounded matrix factorisation aims to find two factor matrices $\matr{W}$ and $\matr{H}$ such that
	\begin{equation}
	\label{mbmf}
	\begin{aligned}
	\matr{V} & \approx \matr{\hat{V}} = \matr{W} \matr{H} \\
	s.t. \quad | \matr{W}_{i:} | & = \bm{r}^{\matr{W}}_{i}, \forall i = 1,2,\cdots,N,\\
	| \matr{H}_{:j} | & = \bm{r}^{\matr{H}}_{j}, \forall j = 1,2,\cdots,M.
	\end{aligned}
	\end{equation}
\end{definition}

According to the above definition, the prediction entry $\matr{\hat{V}}_{ij}$ is now bounded by the product of its two corresponding magnitudes:
\begin{equation}
-\bm{r}^{\matr{W}}_{i} \bm{r}^{\matr{H}}_{j} \leqslant \matr{\hat{V}}_{ij} \leqslant \bm{r}^{\matr{W}}_{i} \bm{r}^{\matr{H}}_{j}.
\end{equation}

In fact, the two magnitude vectors can form a rank one range matrix $\matr{R} = \bm{r}^{\matr{W}} (\bm{r}^{\matr{H}})^{T} \in \mathbb{R}^{N \times M}$. Thus the prediction entry is bounded by its corresponding entry in the range matrix $\matr{R}$:
\begin{equation}
-\matr{R}_{ij} \leqslant \matr{\hat{V}}_{ij} \leqslant \matr{R}_{ij}.
\end{equation}

Notice that all of the entries in the product matrix can be bounded within individually different ranges which are determined by the give constant magnitude vectors, though the ranges are not totally independent.

According to the definition of MBMF, no matter how few observations a user/item has, its corresponding predictions are always bounded as long as the magnitude vectors are set properly. We assume that the magnitude vectors can reflect a relatively stable preference of each user or item for the current factorisation task. So the magnitudes can also be regarded as user/item preferences in recommender systems.

\subsection{Conversion from Constrained to Unconstrained Task}

The magnitude bounded matrix factorisation defined in Eq.(\ref{mbmf}) is equal to a constrained optimisation problem:
\begin{equation}
\label{conopt}
\begin{aligned}
\min_{\matr{W},\matr{H}} \lVert \matr{Z} & \ast (\matr{V} - \matr{W} \matr{H}) \rVert^{2}_{F} \\
s.t. \quad | \matr{W}_{i:} | = & \bm{r}^{\matr{W}}_{i}, \forall i = 1,2,\cdots,N,\\
| \matr{H}_{:j} | = & \bm{r}^{\matr{H}}_{j}, \forall j = 1,2,\cdots,M.
\end{aligned}
\end{equation}
where $\matr{Z}$ is an indicator matrix in which ones denote the existing observations and zeros otherwise. It is not straightforward to work it out because of the magnitude constraints.

Let us now imagine the geometric representation of the magnitude constraints first. Given a two-dimensional variable $\bm{x} = [x_{1},x_{2}]^{T}$ in Euclidean space, setting its magnitude to a constant $|\bm{x}| = r$ limits the possible space of $\bm{x}$ onto a circle with radius $r$. By introducing an angle variable $\phi$, the vector can then be converted into the Polar coordinate system as $\bm{x} = [r\cos\phi,r\sin\phi]^{T}$, which embeds the magnitude constraint into the Polar representation such that calculations on the converted variable do not need to consider the magnitude constraint. Similarly, for higher dimensional variables, the magnitude constraints limit their possible space onto a sphere or hypersphere in the Euclidean system, and conversion to Spherical coordinates will incorporate the magnitudes into their new representations, which ``removes'' the magnitude constraints.

Thus we introduce two angle matrices $\matr{\Phi} \in \mathbb{R}^{N \times (K-1)}$ and $\matr{\Theta} \in \mathbb{R}^{(K-1) \times M}$ to help represent each row of $\matr{W}$ and each column of $\matr{H}$ in Spherical space with the radii set as corresponding magnitudes. A n-dimensional variable $\bm{x}=[x_{1},x_{2},\cdots,x_{n}]^{T}$ can be represented by a radial coordinate $r$ and $n-1$ angular coordinates $\bm{\phi} = [\phi_{1},\phi_{2},\cdots,\phi_{n-1}]^{T}$ where $\phi_{1},\phi_{2},\cdots,\phi_{n-2} \in [0,\pi]$ and $\phi_{n-1} \in [0,2\pi)$:
\begin{equation}
\begin{aligned}
x_{1} & = r \cos\phi_{1}\\
x_{2} & = r \sin\phi_{1}\cos\phi_{2}\\
& \vdots \\
x_{n-1} & = r \sin\phi_{1} \cdots \sin\phi_{n-2}\cos\phi_{n-1}\\
x_{n} & = r \sin\phi_{1} \cdots \sin\phi_{n-2}\sin\phi_{n-1}.
\end{aligned}
\end{equation}

Define auxiliary vectors of functions $\bm{s}(\bm{\phi}) = [s_{1}(\bm{\phi}),s_{2}(\bm{\phi}),$\quad$\cdots,s_{n}(\bm{\phi})]^{T}$ and $\bm{c}(\bm{\phi}) = [c_{1}(\bm{\phi}),c_{2}(\bm{\phi}),\cdots,c_{n}(\bm{\phi})]^{T}$ such that
\begin{equation}
s_{i}(\bm{\phi}) =
\begin{cases}
1, & i = 1 \\
\prod_{j=1}^{i-1} \sin\phi_{j}, & i > 1
\end{cases},
\end{equation}
\begin{equation}
c_{i}(\bm{\phi}) = 
\begin{cases}	
\cos\phi_{i}, & i < n\\
1, & i = n
\end{cases},
\end{equation}
then we have
\begin{equation}
x_{i} = r s_{i}(\bm{\phi}) c_{i}(\bm{\phi}),
\end{equation}
thus
\begin{equation}
\bm{x} = r \bm{s}(\bm{\phi}) \ast \bm{c}(\bm{\phi}).
\end{equation}
Then both of the factor matrices can be converted into the Spherical space:
\begin{equation}
\label{pwi}
\matr{W}_{i:} = \bm{r}^{\matr{W}}_{i} \bm{s}(\matr{\Phi}_{i:}) \ast \bm{c}(\matr{\Phi}_{i:}),
\end{equation}
\begin{equation}
\label{phj}
\matr{H}_{:j} = \bm{r}^{\matr{H}}_{j} \bm{s}(\matr{\Theta}_{:j}) \ast \bm{c}(\matr{\Theta}_{:j}),
\end{equation}

Define auxiliary matrices of functions $\matr{S}(\matr{\Phi})$, $\matr{S}(\matr{\Theta})$ such that
\begin{equation}
\matr{S}_{i:}(\matr{\Phi}) = \bm{s}(\matr{\Phi}_{i:}), \quad \matr{S}_{:j}(\matr{\Theta}) = \bm{s}(\matr{\Theta}_{:j}),
\end{equation}
and $\matr{C}(\matr{\Phi})$, $\matr{C}(\matr{\Theta})$ such that
\begin{equation}
\matr{C}_{i:}(\matr{\Phi}) = \bm{c}(\matr{\Phi}_{i:}), \quad \matr{C}_{:j}(\matr{\Theta}) = \bm{c}(\matr{\Theta}_{:j}),
\end{equation}
then we have
\begin{equation}
\label{pw}
\matr{W} = \bm{r}^{\matr{W}} \odot \matr{S}(\matr{\Phi}) \ast \matr{C}(\matr{\Phi}),
\end{equation}
\begin{equation}
\label{ph}
\matr{H} = \bm{r}^{\matr{H}} \odot \matr{S}(\matr{\Theta}) \ast \matr{C}(\matr{\Theta}),
\end{equation}
where $\odot$ denotes the Khatri-Rao product. Notice that we omit some transpose signs in the formulas from Eq. (\ref{pwi}) to (\ref{ph}) to make them concise and consistent. We refer to this kind of conversion as \textit{Spherical2Cartesian}.

Thus, the constrained optimisation problem in Eq. (\ref{conopt}) is equivalent to the following unconstrained optimisation task:
\begin{equation}
\label{uncopt}
\min_{\matr{\Phi},\matr{\Theta}} \lVert \matr{Z} \ast (\matr{V} - (\bm{r}^{\matr{W}} \odot \matr{S}(\matr{\Phi}) \ast \matr{C}(\matr{\Phi}))(\bm{r}^{\matr{H}} \odot \matr{S}(\matr{\Theta}) \ast \matr{C}(\matr{\Theta}))) \rVert_{F}^{2}.
\end{equation} 

\subsection{Solving Unconstrained Optimisation}

The objective functions defined in Eq. (\ref{conopt}) and Eq. (\ref{uncopt}) both are  non-convex. Fortunately, we found that it can be solved efficiently by the SGD method \cite{BottouCOMPSTAT2010}. According to Eq.(\ref{uncopt}), the objective function is defined as
\begin{equation}
\label{obj}
\mathscr{F} = \lVert \matr{Z} \ast (\matr{V} - (\bm{r}^{\matr{W}} \odot \matr{S}(\matr{\Phi}) \ast \matr{C}(\matr{\Phi}))(\bm{r}^{\matr{H}} \odot \matr{S}(\matr{\Theta}) \ast \matr{C}(\matr{\Theta}))) \rVert_{F}^{2}.
\end{equation}

First, we calculate the partial derivatives of $\mathscr{F}$ with respect to a single element, say $\matr{\Phi}_{ab}$. Here we regard $\matr{W}$ as a function of $\matr{\Phi}$ (so does $\matr{H}$ to $\matr{\Theta}$). Thus by chain rule, we have:
\begin{equation}
\label{dpab}
\frac{\partial\mathscr{F}}{\partial\matr{\Phi}_{ab}} = \sum_{j=1}^{K}\frac{\partial\mathscr{F}}{\partial\matr{W}_{aj}} \frac{\partial\matr{W}_{aj}}{\partial\matr{\Phi}_{ab}} = \frac{\partial\mathscr{F}}{\partial\matr{W}_{a:}} (\frac{\partial\matr{W}_{a:}}{\partial\matr{\Phi}_{ab}})^{T},
\end{equation}
where the summation is due to that $\matr{\Phi}_{ab}$ participants in the representation of all elements of vector $\matr{W}_{a:}$.

The partial derivative of $\mathscr{F}$ with respect to $\matr{W}_{aj}$ is
\begin{equation}
\frac{\partial\mathscr{F}}{\partial\matr{W}_{aj}} = 2((\matr{Z} \ast (\matr{W}\matr{H}))\matr{H}^{T} -(\matr{Z} \ast \matr{V})\matr{H}^{T})_{aj}.
\end{equation}

The partial derivative of $\matr{W}_{aj}$ with respect to $\matr{\Phi}_{ab}$ is
\begin{equation}
\label{dphi}
\frac{\partial\matr{W}_{aj}}{\partial\matr{\Phi}_{ab}} = \bm{r}^{\matr{W}}_{a} \cdot
\begin{cases}
0, & j < b \\
-\prod_{p=1}^{j}\sin\matr{\Phi}_{ap}, & j = b \\
\cos\matr{\Phi}_{ab}\cos\matr{\Phi}_{aj} \prod_{p=1; p \neq b}^{j-1}\sin\matr{\Phi}_{ap}, & b < j < K \\
\cos\matr{\Phi}_{ab} \prod_{p=1; p \neq b}^{j-1}\sin\matr{\Phi}_{ap}, & j = K	
\end{cases}.
\end{equation}

Define auxiliary vectors of functions $\bm{s}(\bm{\Phi}_{a:},b) = [s_{1}(\bm{\Phi}_{a:},b),\cdots,s_{K}(\bm{\Phi}_{a:},b)]^{T}$ and $\bm{c}(\bm{\Phi}_{a:},b) = [c_{1}(\bm{\Phi}_{a:},b),$\quad$\cdots,c_{K}(\bm{\Phi}_{a:},b)]^{T}$ such that
\begin{equation}
s_{j}(\matr{\Phi}_{a:},b) = \prod_{p = 1}^{b-1}\sin\matr{\Phi}_{ap} \cdot
\begin{cases}
0, & j < b \\
-\sin\matr{\Phi}_{aj}, & j = b \\
1, & j = b + 1 \\
\prod_{p=b+1}^{j-1}\sin\matr{\Phi}_{aj}, & b + 1 < j \leqslant K
\end{cases},
\end{equation}
\begin{equation}
c_{j}(\matr{\Phi}_{a:},b) =
\begin{cases}
0, & j < b \\
1, & j = b \\
\cos\matr{\Phi}_{aj} \cos\matr{\Phi}_{ab}, & b < j < K \\
\cos\matr{\Phi}_{ab}, & j = K
\end{cases},
\end{equation}
then we have
\begin{equation}
\frac{\partial\matr{W}_{a:}}{\partial\matr{\Phi}_{ab}} = (\bm{r}_{a}^{\matr{W}} \bm{s}(\bm{\Phi}_{a:},b) \ast \bm{c}(\bm{\Phi}_{a:},b))^{T},
\end{equation}
thus
\begin{equation}
\begin{aligned}
\frac{\partial\mathscr{F}}{\partial\matr{\Phi}_{ab}} = & 2((\matr{Z} \ast (\matr{W}\matr{H}))\matr{H}^{T} -(\matr{Z} \ast \matr{V})\matr{H}^{T})_{a:}\\
& \cdot (\bm{r}_{a}^{\matr{W}} \bm{s}(\bm{\Phi}_{a:},b) \ast \bm{c}(\bm{\Phi}_{a:},b)).
\end{aligned}
\end{equation}

Similarly the derivative of $\mathscr{F}$ with respective to $\matr{\Theta}_{ab}$ is
\begin{equation}
\begin{aligned}
\frac{\partial\mathscr{F}}{\partial\matr{\Theta}_{ab}} = & 2((\matr{W}^{T} (\matr{Z} \ast (\matr{W}\matr{H})) -\matr{W}^{T} (\matr{Z} \ast \matr{V}))_{:b})^{T} \\
& \cdot (\bm{r}_{b}^{\matr{H}} \bm{s}(\bm{\Theta}_{:b},a) \ast \bm{c}(\bm{\Theta}_{:b},a)).
\end{aligned}
\end{equation}

So the updating rules of $\matr{\Phi}$ and $\matr{\Theta}$ using stochastic gradient descent are
\begin{equation}
\label{upab}
\begin{aligned}
\matr{\Phi}_{ab}  \leftarrow  \matr{\Phi}_{ab} - & \lambda^{\matr{\Phi}} ((\matr{Z} \ast (\matr{W}\matr{H}))\matr{H}^{T} -(\matr{Z} \ast \matr{V})\matr{H}^{T})_{a:} \\
& \cdot (\bm{r}_{a}^{\matr{W}} \bm{s}(\bm{\Phi}_{a:},b) \ast \bm{c}(\bm{\Phi}_{a:},b)),
\end{aligned}
\end{equation}
\begin{equation}
\label{utab}
\begin{aligned}
\matr{\Theta}_{ab}  \leftarrow  \matr{\Theta}_{ab} - & \lambda^{\matr{\Theta}} ((\matr{W}^{T} (\matr{Z} \ast (\matr{W}\matr{H})) -\matr{W}^{T} (\matr{Z} \ast \matr{V}))_{:b})^{T} \\
& \cdot (\bm{r}_{b}^{\matr{H}} \bm{s}(\bm{\Theta}_{:b},a) \ast \bm{c}(\bm{\Theta}_{:b},a)),
\end{aligned}
\end{equation}
where $\lambda^{\matr{\Phi}}$ and $\lambda^{\matr{\Theta}}$ are the learning rate parameters.

\subsection{Acceleration for MBMF}

According to our experiments, it is inefficient to individually update each element of $\matr{\Phi}$ and $\matr{\Theta}$ by Eq. (\ref{upab}) and (\ref{utab}). A feasible solution is to update the whole matrices at the same time as long as the derivatives of $\mathscr{F}$ with respect to $\matr{\Phi}$ and $\matr{\Theta}$ can be computed directly.

Referring to Eq. (\ref{dpab}), we know that the derivative of $\mathscr{F}$ with respect to a certain element is a result of a dot product. That is, the derivative of $\mathscr{F}$ with respect to $\matr{\Phi}$ is
\begin{equation}
D^{\matr{\Phi}} = \frac{\partial\mathscr{F}}{\partial\matr{\Phi}}  = 
\left[
\begin{array}{ccc}
\frac{\partial\mathscr{F}}{\partial\matr{W}_{1:}} (\frac{\partial\matr{W}_{1:}}{\partial\matr{\Phi}_{11}})^{T} & \cdots & \frac{\partial\mathscr{F}}{\partial\matr{W}_{1:}} (\frac{\partial\matr{W}_{1:}}{\partial\matr{\Phi}_{1(K-1)}})^{T}\\
\frac{\partial\mathscr{F}}{\partial\matr{W}_{2:}} (\frac{\partial\matr{W}_{2:}}{\partial\matr{\Phi}_{21}})^{T} & \cdots & \frac{\partial\mathscr{F}}{\partial\matr{W}_{2:}} (\frac{\partial\matr{W}_{2:}}{\partial\matr{\Phi}_{2(K-1)}})^{T}\\
\vdots & \ddots & \vdots \\
\frac{\partial\mathscr{F}}{\partial\matr{W}_{N:}} (\frac{\partial\matr{W}_{N:}}{\partial\matr{\Phi}_{n1}})^{T} & \cdots & \frac{\partial\mathscr{F}}{\partial\matr{W}_{N:}} (\frac{\partial\matr{W}_{N:}}{\partial\matr{\Phi}_{n(K-1)}})^{T}\\
\end{array}
\right].
\end{equation}

Notice that the $k^{\text{th}}$ column of $D^{\matr{\Phi}}$ is a vector of dot products of two matrices---$\frac{\partial\mathscr{F}}{\partial\matr{W}}$ and $\frac{\partial\matr{W}}{\partial\matr{\Phi}_{:k}}$ on the row dimension. Thus we can compute $D^{\matr{\Phi}}$ in the following way:

1) calculate the derivative of $\matr{W}$ with respect to $\matr{\Phi}$, which is a $N \times K \times (K - 1)$ three-way tensor (indeed it is a four-way tensor with a singleton dimension);

2) duplicate $\frac{\partial\mathscr{F}}{\partial\matr{W}}$, a $N \times K$ matrix $(K -1)$ times to form the other $N \times K \times (K - 1)$ three-way tensor;

3) $D^{\matr{\Phi}}$ is then the resultant matrix of dot products of the two tensors on the second dimension (dot products of corresponding mode-$2$/row fibres).

The tensor in 2) is straightforward but not in 1). According to Eq. (\ref{dphi}), the derivative of $\matr{W}$ with respect to the $a^{\text{th}}$ row of $\matr{\Phi}$ is a $(K - 1) \times K$ matrix which is nearly upper triangular, and the derivative of $\matr{W}$ with respect to the $b^{\text{th}}$ column of $\matr{\Phi}$ is a $N \times K$ matrix where each row shares the same format. Thus, we can construct the tensor in 1) by calculating the derivatives of $\matr{W}$ with respect to each column of $\matr{\Phi}$ and integrate them together. 

In the algorithm of MBMF, the calculation of the derivative of $\matr{W}$ with respect to $\matr{\Phi}$ is referred as \textit{GradWwrtPhi}.

The derivative of $\mathscr{F}$ with respect to $\matr{\Theta}$ can be obtained in a similar way, and the calculation of the derivative of $\matr{H}$ with respect to $\matr{\Theta}$ is referred as \textit{GradHwrtTheta}. Now the updating rules of $\matr{\Phi}$ and $\matr{\Theta}$ become
\begin{equation}
\label{uwpab}
\matr{\Phi}  \leftarrow  \matr{\Phi} - \lambda^{\matr{\Phi}} D^{\matr{\Phi}},
\end{equation}
\begin{equation}
\label{uwtab}
\matr{\Theta}  \leftarrow  \matr{\Theta} - \lambda^{\matr{\Theta}} D^{\matr{\Theta}}.
\end{equation}

According to the conversion from the Spherical to Cartesian coordinate systems (see \textit{Spherical2Cartesian}), $\matr{W}$ and $\matr{H}$ then can be easily updated with the renewed $\matr{\Phi}$ and $\matr{\Theta}$. The complete algorithm of MBMF is shown in Algorithm \ref{MBMF}.

\begin{algorithm}
	\caption{Magnitude Bounded Matrix Factorisation algorithm (MBMF)}
	\label{MBMF}
	\begin{algorithmic}[1]
		\REQUIRE $\matr{V} \in \mathbb{R}^{N \times M}$ (factorising matrix),\\
		$K \in \mathbb{N}^{+}$ (latent dimension),\\
		$\bm{r}^{\matr{W}}, \bm{r}^{\matr{H}}$ (magnitudes for $\matr{W}$ and $\matr{H}$)\\
		$\lambda_{\matr{\Phi}} \in \mathbb{R}^{+}, \lambda_{\matr{\Theta}} \in \mathbb{R}^{+}$ (learning coefficients)\\
		\ENSURE $\matr{W} \in \mathbb{R}^{N \times K}$ (the left factorised matrix),\\
		$\matr{H} \in \mathbb{R}^{K \times M}$ (the right factorised matrix)\\
		\STATE Initialise $\matr{\Phi}$ and $\matr{\Theta}$ as random angle matrices
		\STATE $\matr{W} \leftarrow \text{Spherical2Cartesian}(\matr{\Phi},\bm{r}^{\matr{W}})$
		\STATE $\matr{H} \leftarrow \text{Spherical2Cartesian}(\matr{\Theta},\bm{r}^{\matr{H}})$
		\WHILE{Terminal conditions not satisfied}
		\STATE $\text{GradW} \leftarrow (\matr{Z} \ast (\matr{W}\matr{H}))\matr{H}^{T} -(\matr{Z} \ast \matr{V})\matr{H}^{T}$
		\STATE $\text{GradW} \leftarrow \text{duplicate}(\text{GradW}, K - 1)$
		\STATE $\text{GradPhi} \leftarrow \textit{GradWwrtPhi}(\matr{\Phi},\bm{r}^{\matr{W}})$		
		\STATE $\matr{\Phi} \leftarrow \matr{\Phi} - \lambda^{\matr{\Phi}}\cdot \text{dot}(\text{GradW},\text{GradPhi},2)$
		\STATE $\text{GradH} \leftarrow \matr{W}^{T} (\matr{Z} \ast (\matr{W}\matr{H})) -\matr{W}^{T} (\matr{Z} \ast \matr{V})$
		\STATE $\text{GradH} \leftarrow \text{duplicate}(\text{GradH}, K - 1)$
		\STATE $\text{GradTheta} \leftarrow \textit{GradHwrtTheta}(\matr{\Theta},\bm{r}^{\matr{H}})$
		\STATE $\matr{\Theta} \leftarrow \matr{\Theta}- \lambda^{\matr{\Theta}}\cdot \text{dot}(\text{GradH},\text{GradTheta},1)$
		\STATE $\matr{W} \leftarrow \textit{Spherical2Cartesian}(\matr{\Phi},\bm{r}^{\matr{W}})$
		\STATE $\matr{H} \leftarrow \textit{Spherical2Cartesian}(\matr{\Theta},\bm{r}^{\matr{H}})$
		\ENDWHILE
	\end{algorithmic}
\end{algorithm}

\subsection{Pre-processing \& Choice of Magnitudes}

MBMF is an algorithm which bounds each entry of the product matrix in a centrosymmetric range with respect to the origin point zero. We denote the proposed algorithm applied on centred data as MBMF-c. However, MBMF is also workable when it only limits the upper bound of the entries, which requires the data to be non-negative. We refer to this kind of algorithm applied on non-negative data as MBMF-n. Because of this feature in MBMF, we discuss three different types of data and the choice of magnitudes for each type in this section. We also introduce a feasible way to incorporate historical data into magnitudes which has been used in our experiments.

There are three different types of data with respect to bounds in recommender systems: (i) the data is bounded within a single predefined range $[r_{\text{min}}, r_{\text{max}}]$; (ii) the data only has one upper or lower bound (e.g. non-negative $[0, \infty)$); (iii) the data has no bounds.

\textit{Type (i) Data}. This type of data is most commonly seen in classic recommender system studies. In this case, for MBMF-c, the observations can be simply centred by
\begin{equation}
\label{cen1}
\matr{V}_{ij} = \matr{V}_{ij} - \frac{r_{\text{min}} + r_{\text{max}}}{2}, \forall i,j, \matr{Z}_{ij} = 1,
\end{equation}
and the magnitudes can be set as any values satisfying the following equation:
\begin{equation}
\label{magc}
\bm{r}^{\matr{W}}_{i} \bm{r}^{\matr{H}}_{j} = \frac{r_{\text{max}} - r_{\text{min}}}{2},\forall i,j.
\end{equation}

For MBMF-n, the non-negativity can be achieved by
\begin{equation}
\label{non}
\matr{V}_{ij} = \matr{V}_{ij} - r_{\text{min}}, \forall i,j, \matr{Z}_{ij} = 1,
\end{equation}
or keeping them unchanged if $r_{\text{min}} \geqslant 0$. Meanwhile, the magnitudes can be set as:
\begin{equation}
\label{magn}
\bm{r}^{\matr{W}}_{i} \bm{r}^{\matr{H}}_{j} = r_{\text{max}},\forall i,j
\end{equation}

Take the movie ratings in Movielens dataset as an example. The predefined rating range is $[1,5]$ and all the observations are integers. For MBMF-c, centring by Eq. (\ref{cen1}) turns $1,2,3,4,5$ to $-2,-1,0,1,2$ and the magnitudes can be set all $\sqrt{2}$. Then the entries in the product are bounded within $[-2, 2]$. For MBMF-n, the observations are already non-negative, thus the magnitudes can simply be set all $\sqrt{5}$ which bound all products to be within $[-5, 5]$. In our experiments, it is very rare that entries in product matrix are smaller than the predefined lower bound when applying MBMF-n. We only found exceptions when most observations were close to the lower bound.

\textit{Type (ii) Data}. Non-negative observations with no upper bound are the most typical ones of type (ii) data. In this case, non-negativity is natural but centring is no longer straightforward. However, it can still be done if the magnitudes are available. They can be obtained from additional source of information, such as historical ratings, user/item profiles and social connections. Once the magnitudes of all users and items are determined, under the consideration of non-negativity, the centring formula for known observations is as follows:
\begin{equation}
\label{cen2}
\matr{V}_{ij} = \matr{V}_{ij} - \bm{r}^{\matr{W}}_{i} \bm{r}^{\matr{H}}_{j}, \forall i,j, \matr{Z}_{ij} = 1.
\end{equation}

Note that there might be contradictions when centring data in this scenario. For example, if the original observation is $5$ and the corresponding magnitudes calculated from meta information are $1$ and $2$, then applying Eq. (\ref{cen2}) will set $3$ as the centred entry, which is not within the centred range $[-2,2]$. Basically, this contradiction happens when $\matr{V}_{ij} > 2\bm{r}^{\matr{W}}_{i} \bm{r}^{\matr{H}}_{j}$.

Solutions for the contradiction include regarding the observation as an outlier (remove it from the dataset), modifying the magnitudes to cover the observation, and trying to use different sources of meta information.

\begin{table*}[!ht]
	\caption{Running Time (s) on Different Scales of Matrix.}
	\label{complexity}
	\begin{center}
		\begin{tabular}{|c|c|c|c||C|C|C|C|}
			\hline
			N & M & K & density & BMF & BMC-ADMM & Bounded-SVD & MBMF-n \\
			\hline
			$20$ & $50$ & 10 & 1 & \textbf{0.1497} & 2.2388 & 0.2631 & 0.3018 \\
			$100$ & $100$ & 10 & 1 & \textbf{0.5007} & 8.8932 & 2.6718 & 0.7857 \\
			$200$ & $500$ & 10 & 1 & 5.7100 & 4.4400 & 23.291 & \textbf{3.0845} \\
			$1,000$ & $1,000$ & 10 & 1 & 313.75 & \textbf{14.908} & 201.71 & 18.293 \\
			$2,000$ & $5,000$ & 10 & 1 & 4,753.4 & 128.73 & 2,101.3 & \textbf{91.020} \\
			\hline
			\multicolumn{4}{|c||}{Avg.} & 1,014.7 & 31.841 & 465.84 & \textbf{22.697} \\
			\hline\hline
			$100,000$ & $100,000$ & 10 & 0.001 & - & 34,025 & - & \textbf{5,728.3} \\
			$200,000$ & $500,000$ & 10 & 0.0002 & - & 442,250 & - & \textbf{12,251} \\
			\hline
			\multicolumn{4}{|c||}{Avg.} & - & 238,138 & - & \textbf{8989.7} \\
			\hline
		\end{tabular}
	\end{center}
\end{table*}

\textit{Type (iii) Data}. As for data with no bounds, we can only assume that $r_{\text{max}} = \max{\matr{V}}, r_{\text{min}} = \min{\matr{V}}$. Applying Eq. (\ref{non}) will make all the observations non-negative, while centring is impossible in this case because there is no reference. However, MBMF-c is still workable as long as the magnitudes obtained can cover all the observations in their corresponding ranges.

\textit{Incorporating Historical Data}. Using historical ratings to get magnitudes is feasible and practical since most of the real world recommender systems have historical records. Here, we introduce what we have done to obtain good magnitudes based on historical data in our implementation.

Assume that $\matr{V}$ is the current factorising matrix and $\matr{V}^{\prime}$ is the corresponding historical matrix. Consider the recommendation for user $i$. A possible way to calculate the magnitude of this user is
\begin{equation}
\label{mag2}
\bm{r}_{i}^{\matr{W}} = \sqrt{\mathbb{E}(\matr{V}^{\prime}_{i:}) + \sigma(\matr{V}^{\prime}_{i:})},
\end{equation}
where $\mathbb{E}(\cdot)$ is the mean value and $\sigma(\cdot)$ is the standard deviation. This magnitude is based on the average rating and allows possible variations. However, there might be cases in which a user has very few historical records, and applying Eq. (\ref{mag2}) to calculate the magnitude in these cases can cause over-fitting issues. For example, if user $i$ only has rated a movie $1$ out of $5$ in the past, Eq. (\ref{mag2}) gives $\bm{r}_{i}^{\matr{W}} = 1$ since the mean is $1$ and the standard deviation is $0$, which means this user will be super strict for all of its future ratings. Thus, we introduce the global influence for individuals to mitigate such issues:
\begin{equation}
\label{mag3}
\bm{r}_{i}^{\matr{W}} = \omega_{i} \sqrt{\mathbb{E}(\matr{V}^{\prime}_{i:}) + \sigma(\matr{V}^{\prime}_{i:})} + (1 - \omega_{i}) \sqrt{\mathbb{E}(\matr{V}^{\prime}) + \sigma(\matr{V}^{\prime})},
\end{equation}
where $\omega_{i}$ is the weight of how much impact in the magnitude is from the user itself. In our implementation, we use the following formula to determine the value of $\omega_{i}$:
\begin{equation}
\label{omega}
\omega_{i} = 
\begin{cases}
0, & \text{user $i$ has no historical data} \\
\min(\sum\matr{Z}^{\prime}_{i:} / (\rho M), 1), & \text{otherwise}
\end{cases}
\end{equation}
where $\matr{Z}^{\prime}$ is the indicator matrix for historical data, $\rho \in (0, 1]$ is a consistent percentage denoting how much historical data a user needs to completely represent its own preferences, and $M$ is the number of items. More specifically, if a user has no records (i.e. $\omega_{i} = 0$), MBMF uses the global impact from all other users to define its magnitude. If a user has rated more than $\rho$ of all the items in the past (i.e. $\omega_{i} = 1$), MBMF will fully use its historical ratings to determine its magnitude; the more historical data a user has, the more weight will be imposed on the impact from its own, and vice versa.

\subsection{Parameter Setting}
The parameters of MBMF are the learning coefficients $\lambda_{\matr{\Phi}}$ and $\lambda_{\matr{\Theta}}$ which are introduced by the SGD method. Among literature about SGD, the learning coefficients or the step sizes can be set in three different ways: (1) set the steps the value that produce the best performance after several tries; (2) change the steps dynamically during the updating procedure by checking the change of the value of objective function: if the function value decreases, make the steps a bit bigger (e.g. multiplied by $1.1$), otherwise shrink them (e.g. divided by $2$) and roll back; (3) apply a line search method (e.g. bisection search or backtracking search with Wolfe conditions) to determine more reasonable steps. We used the dynamic setting for parameters in our implementation, and the initial values for both parameters are set at $0.1$.

\subsection{Complexity Analysis}

We analyse the computation complexity of the proposed MBMF algorithm both theoretically and numerically.

For each iteration, MBMF first calculates the partial derivatives of $\mathscr{F}$ with respect to $\matr{W}$ and $\matr{H}$, which costs $2NMK + 2(N+M)K^{2}$ addition and $2NMK + 2(N+M)K^{2} + (N + M)K$ multiplication operations; then it computes the partial derivatives of $\matr{W}$ and $\matr{H}$ with respect to $\matr{\Phi}$ and $\matr{\Theta}$, and requires $2(N+M)(K-1)$ trigonometric operations plus $2(N+M)(K-2)(K-1)$ times multiplication; it also operates $2(N+M)(K-2)$ multiplication to transform $\matr{\Phi}$ and $\matr{\Theta}$ to $\matr{W}$ and $\matr{H}$. Thus the overall complexity of MBMF is $O(NMK)$, which is the same as the traditional matrix factorisation approach.

We conduct experiments to show the time consumed of MBMF on $7$ different scales of matrices against other bounding algorithms BMF, BMC-ADMM and Bounded-SVD. The matrix scale varies from $10^{3}$ to $10^{11}$. The experiments are divided into two groups. One is conducted on small and full matrices with all $7$ algorithms (repeated $5$ times), and the other is conducted on large and sparse matrices (run once) without BMF and Bounded-SVD since they are extremely time consuming. All complexity experiments are conducted on a computer with i7-6900K CPU, 16G memory and Windows 10. The average results are presented in TABLE \ref{complexity}. 

The results show that, (i) BMF and Bounded-SVD are quite fast when the matrix is small, but their running time increases when the matrix size becomes larger. Since the third case study, they start to fall far behind to the other bounding algorithms, and for the group of large scale matrices, they are even incapable of converging in a reasonable time. (ii) BMF-ADMM is not the fastest algorithm on small matrices, but it can still be applied to large scale instances, despite its long running time. (iii) MBMF-n is the overall fastest algorithm not only on small matrices, but also on large scale ones. This demonstrates that MBMF is the most efficient bounding algorithm and has great advantages when applied on large-scale recommender systems.

\section{Experiments}

We explore prediction variation of algorithms on synthetic datasets and compare the recommendation performance against related algorithms on several real datasets including Jester \cite{GoldbergIR2001}, BookCrossing \cite{ZieglerWWW2005}, and Tianchi Mobile Record datasets. The statistics of these datasets after preprocessing (i.e. removing invalid records) are presented in TABLE \ref{st}. The baseline algorithms we chose for comparison with the proposed MBMF-n and MBMF-c are: 
\begin{itemize}
	\item {
		\textit{Matrix Factorisation for recommender systems (MF)} \cite{KorenC2009}. The conventional approach used for recommender systems.
	}
	\item {
		\textit{Non-negative Matrix Factorisation algorithms (NMF)} \cite{LeeNIPS2001}. Using ``multiplicative rules'' instead of addition formula in the SGD, NMF becomes popular when dealing with non-negative observations.
	}
	\item {
		\textit{Feature-wise updated Cyclic Coordinate Descent method (CCD++)} \cite{YuICDM2012}. It is a new and fast non-bounding approach to handle large-scale datasets.
	}
	\item {
		\textit{Bounded Matrix Factorisation (BMF)} \cite{KannanKIS2014}. This is the first algorithm studying bounding issues in matrix factorisation for recommender systems.
	}
	\item {
		\textit{Bounded Matrix Completion in Alternating Direction of Multiplier Method (BMC-ADMM)} \cite{FangIJCAI2017}. It solved the convergence issue in BMF with the help of ADMM framework.
	}
	\item {
		\textit{Bounded Singular Value Decomposition (Bounded-SVD)} \cite{LeICEITES2015}. This approach tries to bound factorising entries by imposing exponential penalties.
	}
\end{itemize}

\begin{table}[!htbp]
	\caption{Statistics of Datasets used for Experiments}
	\begin{center}
		\label{st}
		\begin{tabular}{|c|c|c|c|c|c|}
			\hline
			dataset & users & items & density & range & ordinal\\
			\hline
			Synthetic & 500 & 500 & 0.2 & [0,10] & N\\
			\hline
			Jester & 73,421 & 100 & 0.5634 & [-10,10] & N\\
			BookCrossing & 92,184 & 270,169 & 0.00004 & [1,10] & Y\\
			Ali Mob & 10,000 & 8,916 & 0.0101 & [1,6853] & Y\\
			\hline
		\end{tabular}
	\end{center}
\end{table} 

Three metrics are used to evaluate the performance of recommendations:

(1) \textit{Root Mean Square Error (RMSE)} \cite{BarnstonWF1992}, which evaluates difference between the recovered ratings and their corresponding original ratings:
\begin{equation}
\text{RMSE} = (\sum_{ij}\matr{M}_{ij})^{-1/2}\lVert \matr{M} \ast (\matr{V} - \matr{W}^{\ast}\matr{H}^{\ast}) \rVert_{F},
\end{equation}
where $\matr{W}^{\ast}$ and $\matr{H}^{\ast}$ are the final factorised matrices, and $\matr{M}$ is a binary matrix in which one denotes the chosen entry for cross validation.

(2) \textit{Mean Absolute Error (MAE)} \cite{WillmottCR2005}, which evaluates the absolute difference between the recovered ratings and their corresponding original ratings:
\begin{equation}
\text{MAE} = \frac{1}{\sum_{ij}\matr{M}_{ij}} | \matr{M} \ast (\matr{V} - \matr{W}^{\ast}\matr{H}^{\ast}) |.
\end{equation}

(3) \textit{F1 score} \cite{PowersBP2011}, which is a percentage value that evaluates how much proportion of correct recommendations is with respect to users:
\begin{equation}
\text{F1} = \frac{2 \times \text{Precision} \times \text{Recall}}{\text{Precision} + \text{Recall}}.
\end{equation}
In our experiments, for each user we use its average rating as a threshold, and ratings greater than it imply that the corresponding items are recommended. Before and after factorisation, we obtain two binary recommendation vectors which can be used to calculate the F1 score.

For RMSE and MAE, lower is better, while for F1 Score, higher is better.

The maximal iteration times for all of the algorithms are set at $500$. An early termination condition is also applied: if the decrease of the objective function value is smaller than $10^{-5}$ for $10$ consecutive iterations, the algorithm is regarded as converged.

The source codes of MBMF as well as all the preprocessed datasets we used for experiments can be downloaded at:  https://github.com/shawn-jiang/MBMF.

\subsection{Prediction Variation}

\begin{figure*}[htbp]
	\begin{subfigure}[t]{0.47\linewidth}
		\begin{center}
			\includegraphics[width=\textwidth]{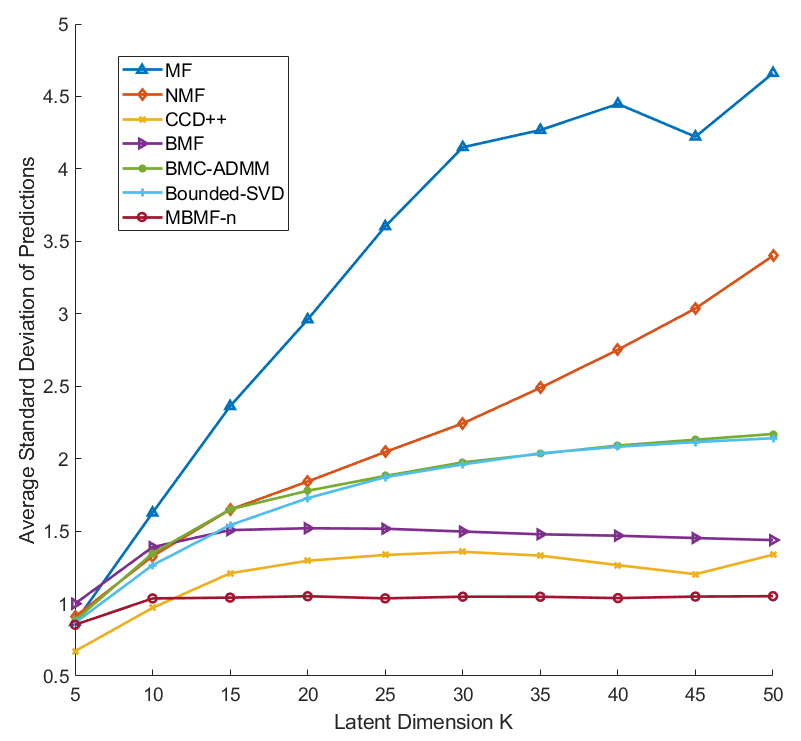}
			\caption{$\text{ave}(\sigma)$ of predictions.}
		\end{center}
	\end{subfigure}
	\hspace{5ex}
	\begin{subfigure}[t]{0.47\linewidth}
		\begin{center}
			\includegraphics[width=\textwidth]{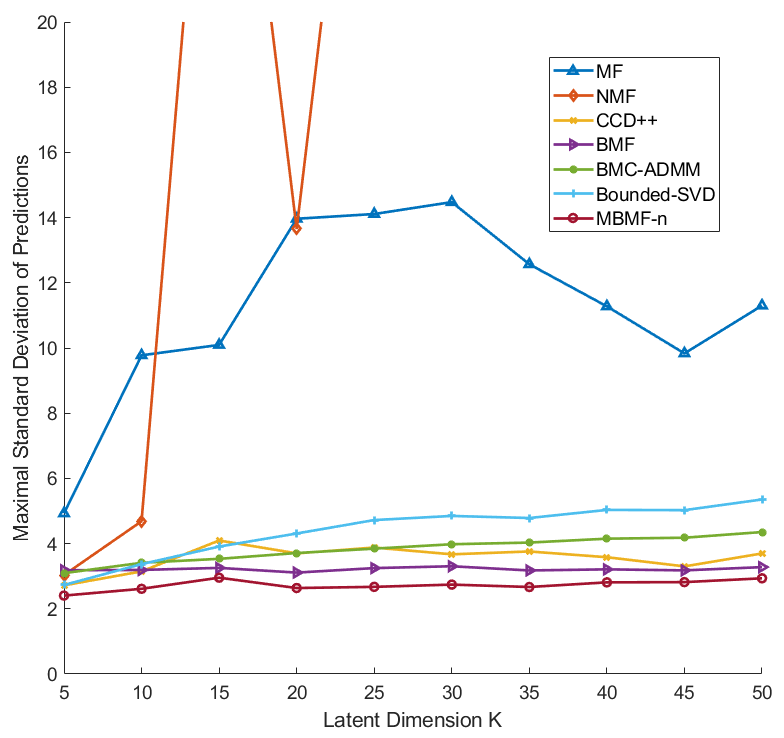}
			\caption{$\text{max}(\sigma)$ of predictions.}
		\end{center}
	\end{subfigure}
	\caption{Prediction variation on Synthetic dataset. \label{prediction}}
\end{figure*}

We conduct experiments on synthetic datasets to explore the prediction variation of algorithms. In this experiment, $K$ varies from $5$ to $50$ with step $5$. For each $K$, we adopt the following experimental procedure:
\begin{itemize}
	\item {
		Randomly generate a factorising matrix $\matr{V}$ of size $500 \times 500$ and scale its entries to $[0, 10]$;
	}
	\item{
		Remove $80\%$ of the values (set them to $0$) in $\matr{V}$ randomly while making sure the remaining part does not have zero rows/columns;
	}
	\item{
		Run each of the algorithms on the remaining matrix $10$ times, and record their recovered matrices;
	}
	\item{
		Calculate the standard deviation ($\sigma$) on missing entries.
	}
\end{itemize} 
The average and maximal standard deviation over predictions is presented in Fig. \ref{prediction}. The lower the standard deviation is, the more stable the predictions are. Algorithms with stable predictions are more operational on large datasets.

According to Fig. \ref{prediction} (note that in Fig. \ref{prediction} (b) the maximal $\sigma$ of NMF is greater than the limit of the y-axis in many cases, since we would like to show more details for other algorithms), we can see that four bounding algorithms (BMF, BMC-ADMM, Bounded-SVD and MBMF-n) all successfully reduced the prediction variation compared with the traditional matrix factorisation methods (MF and NMF) which have no bounding conditions. MBMF-n achieves both the lowest average $\sigma$ and the lowest maximal $\sigma$ in all cases, followed by CCD++ (the performance of CCD++ is impressive although it has no bounding constraints), BMF and Bounded-SVD. This experiment demonstrates that MBMF is so far the best bounding algorithm in terms of reducing prediction variation, and can be easily applied on real recommender systems: it does not require repeated runs.

\subsection{Performance on Real Recommender Systems}

\begin{table*}[!ht]
	\setlength{\abovecaptionskip}{-2pt} 
	\setlength{\belowcaptionskip}{0pt}
	\caption{Recommendation Performance on Real Datasets}
	\begin{center}
		\label{recommender}
		\begin{tabular}{|c|c||D|D|D|D||D|D|D|D||D|D|D|D|}
			\hline
			& & \multicolumn{4}{c||}{RMSE} & \multicolumn{4}{c||}{MAE} & \multicolumn{4}{c|}{F1 Score}\\
			\cline{2-14}
			& K & 10 & 20 & 50 & Avg. & 10 & 20 & 50 & Avg. & 10 & 20 & 50 & Avg.\\
			\hline
			\multirow{7}{*}{Jester} & MF & 5.6107 & 4.9242 & 4.6770 & 5.0710 & 4.2949 & 3.8589 & 3.7919 & 3.9820 & 60.54 & 62.26 & 64.35 & 62.38 \\
			& NMF & 333.18 & 98.349 & 10.161 & 147.20 & 9.3592 & 9.3339 & 5.6652 & 8.1190 & 57.96 & 56.02 & 58.07 & 57.35 \\
			& CCD++ & 5.6643 & 5.5892 & 5.1609 & 5.4710 & 4.2840 & 4.2523 & 3.9493 & 4.1620 & 60.40 & 60.08 & 60.46 & 60.31 \\
			& BMF & 5.4271 & 5.5027 & 5.3717 & 5.4340 & 4.2851 & 4.3581 & 4.2744 & 4.3060 & 58.62 & 58.96 & 58.79 & 58.79 \\	
			& BMC-ADMM & 4.9162 & 4.9477 & 4.7541 & 4.8730 & 3.8149 & 3.8807 & 3.7961 & 3.8310 & 60.52 & 57.42 & 55.22 & 57.72 \\	
			& Bounded-SVD & 6.3798 & 6.7069 & 5.7458 & 6.2780 & 4.7758 & 5.1434 & 4.5270 & 4.8150 & 59.90 & 58.32 & 58.82 & 59.01 \\
			& MBMF-n & \textbf{4.6474} & \textbf{4.6399} & \textbf{4.6344} & \textbf{4.6410} & \textbf{3.7467} & \textbf{3.7473} & \textbf{3.7404} & \textbf{3.7450} & \textbf{65.58} & \textbf{65.71} & \textbf{65.60} & \textbf{65.63} \\
			& MBMF-c & 4.6816 & 4.6695 & 4.6773 & 4.6760 & 3.8045 & 3.8014 & 3.8039 & 3.8030 & 64.08 & 64.47 & 64.16 & 64.24 \\
			\hline
			\multirow{7}{*}{\shortstack{Book-\\crossing}} &	MF & 1.7566 & 1.3539 & 1.6726 & 1.5940 & 1.3164 & 0.9979 & 1.2578 & 1.1910 & \textbf{87.25} & 86.85 & 77.31 & 83.80 \\
			& NMF & 1.8827 & 1.6211 & 1.4146 & 1.6390 & 1.2889 & 1.1479 & 0.9779 & 1.1380 & 82.68 & 83.98 & 86.43 & 84.36 \\
			& CCD++ & 1.3137 & 1.3191 & 1.3131 & 1.3150 & 0.8713 & 0.8753 & 0.8699 & 0.8722 & 87.05 & 87.07 & \textbf{87.12} & \textbf{87.08} \\
			& BMF & 1.3409 & 1.2863 & 1.2990 & 1.3090 & 0.9329 & 0.8563 & 0.8653 & 0.8848 & 86.29 & 86.29 & 86.06 & 86.21 \\
			& BMC-ADMM & 3.3757 & 3.0368 & 2.8886 & 3.1000 & 3.0033 & 2.6781 & 2.5543 & 2.7450 & 81.93 & 78.93 & 67.27 & 76.04 \\		
			& Bounded-SVD & 1.6221 & 1.3541 & 1.6021 & 1.5260 & 1.1873 & 0.9712 & 1.2017 & 1.1200 & 87.02 & \textbf{87.47} & 77.94 & 84.14 \\
			& MBMF-n & 1.2598 & 1.2639 & 1.2628 & 1.2620 & 0.8503 & 0.8534 & 0.8526 & 0.8521 & 86.33 & 86.30 & 86.32 & 86.32 \\
			& MBMF-c & \textbf{1.1783} & \textbf{1.1846} & \textbf{1.1813} & \textbf{1.1810} & \textbf{0.7810} & \textbf{0.7849} & \textbf{0.7817} & \textbf{0.7825} & 85.56 & 85.64 & 85.63 & 85.61 \\
			\hline
			\multirow{6}{*}{\shortstack{Ali \\ Mob}} & MF & 115.93 & 95.286 & 61.138 & 90.784 & 29.930 & 34.871 & 29.975 & 31.592 & 38.99 & 38.22 & 38.76 & 38.66 \\
			& NMF & 235511 & 4540.0 & 11238 & 131098 & 2686.0 & 959.17 & 1223.1 & 1622.8 & 35.90 & 34.29 & 34.77 & 34.99 \\
			& CCD++ & 63.087 & 59.334 & 56.933 & 59.785 & 22.667 & 25.103 & 26.128 & 24.633 & 39.58 & 38.66 & 38.29 & 38.84 \\
			& BMF & 549.66 & 455.34 & 380.98 & 461.99 & 507.57 & 428.79 & 362.07 & 432.81 & 32.07 & 32.28 & 31.69 & 32.01 \\	
			& BMC-ADMM & 319.65 & 392.39 & 593.43 & 435.16 & 157.90 & 205.80 & 358.47 & 240.73 & 17.70 & 18.39 & 22.85 & 19.65 \\	
			& Bounded-SVD & 43.381 & \textbf{42.810} & \textbf{41.766} & \textbf{42.652} & 15.827 & 15.953 & 16.438 & 16.073 & 41.83 & 42.41 & 42.68 & 42.31 \\
			& MBMF-n & \textbf{42.540} & 42.903 & 42.999 & 42.814 & 18.963 & 18.948 & 18.896 & 18.935 & 39.64 & 39.70 & 39.72 & 39.69 \\
			& MBMF-c & 43.273 & 43.671 & 43.881 & 43.608 & \textbf{14.405} & \textbf{14.329} & \textbf{14.400} & \textbf{14.378} & \textbf{43.47} & \textbf{43.51} & \textbf{43.56} & \textbf{43.51} \\
			\hline
		\end{tabular}
	\end{center}
\end{table*}

As for the experiments on real recommendation datasets, we first randomly divided each of them into half historical data and half present data. The historical data can have zero rows/columns while the present data cannot. Then in each run we randomly picked $10\%$ entries from the present data as a validation fold. The latent dimension $K$ varies in $10, 20, 50$. We ran $5$ folds for each $K$ and report the average results in TABLE \ref{recommender}.

%

\textit{\textbf{Jester}} is a free online dataset containing about 4.1 million continuous ratings ($[-10, 10]$) of $100$ jokes from $73,421$ users collected between April 1999 and May 2003. Due to the high density, the continuity in ratings and the centrosymmetric bounds, which are rarely seen in other recommender systems, this dataset becomes popular to test recommendation algorithms.

From TABLE \ref{recommender}, we can see that the performance of MBMF-n on the Jester dataset is the best among all methods, which is $4.6410$ RMSE, $3.7450$ MAE and $65.63\%$ F1 Score, followed by MBMF-c with $4.6760$ RMSE, $3.8030$ MAE and $64.24\%$ F1 Score. The improvement of MBMF is massive on this dataset. We believe this is because the magnitudes obtained from the historical ratings can represent users and jokes very well.

\textit{\textbf{Bookcrossing}} is a large online dataset which contains 278,858 users with 1,149,780 ratings (explicit / implicit) about 271,379 books. However, neither all users nor all books have corresponding ratings in this dataset. From the ratings file, we constructed a rating matrix with $92,184$ rows (users) and $270,169$ columns (books).  

We only did one experiment using BMF for each $K$ on this dataset since it is too time consuming (the experiment with $K$ set $50$ took about $30$ days). From the table of results, MBMF-c achieves the best RMSE ($1.1810$) and MAE ($0.7825$) in all cases, while CCD++ obtains the highest average F1 Score ($87.08\%$) followed by MBMF-n ($86.32\%$).

\textit{\textbf{Ali Mob}} is a public recommendation dataset provided by Tianchi platform. It recorded the online shopping behaviours of users on Alibaba's M-Commerce platforms in 2014. The dataset consists of $10,000$ users and $2,876,947$ commodities in $8,916$ categories. $12,256,906$ valid records were collected in this dataset which contained users' behaviours (click, collect, add-to-cart and payment) towards commodities. We preprocessed this dataset as following:
\begin{itemize}
	\item {
		Count the number of each behaviour ($b_{1},b_{2},b_{3},b_{4}$) for each user ($u$) on each category of commodities ($c$);
	}
	\item {
		Calculate the interest points ($p$) of user $u$ on category $c$:
		\begin{equation}
		\begin{small}
		p = w_{1}b_{1} + w_{2}b_{2} + w_{3}b_{3} + w_{4}b_{4},
		\end{small}
		\end{equation}
		where $w_{1},w_{2},w_{3},w_{4}$ are behaviour weights. We set $w_{1} = 1, w_{2} = 2, w_{3} = 3, w_{4} = 5$ in our experiments according to the increase of interest from ``click'' to ``payment''.
	}
	\item {
		 Construct the data matrix with interest points whose rows denote users and columns denote categories.
	}
\end{itemize} 

This dataset does not have a predefined pair of bounds for entries. However, all values in the factorising matrix are non-negative. Thus, it is type (ii) data. Since the range of observations varies among users/items, we use the maximal value and minimal value as the upper and lower bounds respectively for algorithms (i.e. BMF, BMC-ADMM and Bounded-SVD) which can only handle one single pair of bounds. The parameter $\lambda$ for BMC-ADMM is set $1$ and $\eta$ for Bounded-SVD is set $0.05$.

As shown in TABLE \ref{recommender}, MBMF-c achieves the best performance over MAE ($14.378$ in average) and F1 score ($43.51\%$ in average) in all cases, while Bounded-SVD obtains the lowest RMSE ($42.652$ in average) followed by MBMF-n ($42.814$ in average).

\subsection{Discussion}

From the experimental results, we found an interesting and noteworthy phenomenon: the performance of MBMF does not change much when $K$ varies, while other methods obtain their best results with different $K$ when applied on different datasets. This observation reveals the high stability, a unique feature of MBMF which other methods do not have. It implies that the selection of latent dimension $K$ has little impact on the performance of MBMF. This is very useful when applying MBMF on large scale datasets because it means we do not have to repeat experiments to determine the best suitable $K$ which is usually required by existing methods. The reason of the high stability in MBMF is the magnitude constraint: no matter what $K$'s value is, the magnitude of row/column in the factorised matrices is the same.

To some degree, however, the magnitude constraints are quite strict as well: since all the rows/columns in the factorised matrices have to conform to these magnitude constraints, there is not much space for the SGD method to approximate the product of factorised matrices to be as close to the factorising matrix as in other algorithms. In our future work, we will relax the magnitude constraints from equalities to inequalities (i.e. $| \matr{W}_{i:} | \leqslant \bm{r}^{\matr{W}}_{i}$), which will make MBMF more flexible but harder to solve as well. We believe that imposing magnitude constraints is an effective way to solve bounded matrix factorisation.

\section{Conclusions}
In this paper, we proposed a novel matrix factorisation algorithm for recommender systems called MBMF. It is a bounded algorithm which reduces the fluctuation of predictions especially for large and sparse datasets. The magnitude constraints allow MBMF to have individually different bounds for each entry and make the algorithm capable of dealing with data that has no fixed bounds. The conversion of coordinates used in the method of MBMF turns the magnitude constrained optimisation into an equivalent unconstrained one by setting the radii as corresponding magnitudes. As for the unconstrained task, we apply a stochastic gradient descent method to solve it efficiently. Compared with the unconstrained MF, NMF and CCD++ as well as the-state-of-art bounding methods BMF, BMC-ADMM and Bounded-SVD, MBMF achieved superior performance over all evaluation metrics on both synthetic datasets and real recommendation datasets.

\ifCLASSOPTIONcaptionsoff
  \newpage
\fi



%
\bibliographystyle{IEEEtran}
\bibliography{IEEEtran}

%

\begin{IEEEbiography}[{\includegraphics[width=1in,height=1.25in,clip,keepaspectratio]{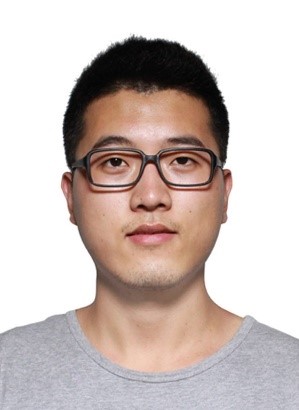}}]{Shuai Jiang}
	received the bachelor’s degree in computer science and technology from Beijing Institute of Technology, Beijing, China, in 2013. Currently he is working towards the dual doctoral degree in both Beijing Institute of Technology and University of Technology Sydney. His main interests include machine learning, optimisation and data analytics.
\end{IEEEbiography}

\begin{IEEEbiography}[{\includegraphics[width=1in,height=1.25in,clip,keepaspectratio]{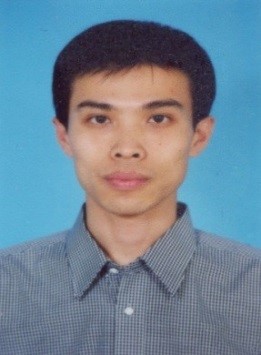}}]{Kan Li}
	is currently a Professor in the School of Computer at Beijing Institute of Technology. He has published over 50 technical papers in peer-reviewed journals and conference proceedings. His research interests include machine learning and pattern recognition.
\end{IEEEbiography}

\begin{IEEEbiography}[{\includegraphics[width=1in,height=1.25in,clip,keepaspectratio]{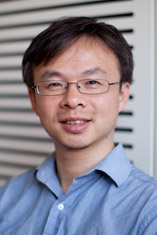}}]{Richard Yi Da Xu}
	received the B.Eng. degree in computer engineering from the University of New South Wales, Sydney, NSW, Australia, in 2001, and the Ph.D. degree in computer sciences from the University of Technology at Sydney (UTS), Sydney, NSW, Australia, in 2006. He is currently an Associate Professor of School of Electrical and Data Engineering, UTS. His current research interests include machine learning, deep learning, data analytics and computer vision.
\end{IEEEbiography}




\end{document}